\documentclass[conference]{IEEEtran}
\IEEEoverridecommandlockouts

\usepackage{cite}
\usepackage{amsmath,amssymb,amsfonts}
\usepackage{algorithmic}
\usepackage{graphicx}
\usepackage{textcomp}
\usepackage{xcolor}
\usepackage{xurl}
\usepackage{float}
\def\BibTeX{{\rm B\kern-.05em{\sc i\kern-.025em b}\kern-.08em
    T\kern-.1667em\lower.7ex\hbox{E}\kern-.125emX}}
\usepackage{balance}

\begin{document}

\title{Product Digital Twin Supporting End-of-life Phase of Electric Vehicle Batteries Utilizing Product--Process--Resource Asset Network}

\author{
    \IEEEauthorblockN{
        Sára Strakošová\IEEEauthorrefmark{1}\IEEEauthorrefmark{3},
        Petr Novák\IEEEauthorrefmark{3},
        Petr Kadera\IEEEauthorrefmark{3}
    }
    \IEEEauthorblockA{
        \IEEEauthorrefmark{1}\textit{Faculty of Mechanical Engineering, Czech Technical University in Prague}, Prague, Czech Republic\\
        sara.strakosova@fs.cvut.cz \\
        \IEEEauthorrefmark{3}\textit{Czech Institute of Informatics, Robotics and Cybernetics, Czech Technical University in Prague}, Prague, Czech Republic\\
        \{first\_name.last\_name\}@cvut.cz
    }
}

\newcommand\textline[4][t]{
  \par\smallskip\noindent\parbox[#1]{1\textwidth}{\raggedright#2}%
  \parbox[#1]{.2\textwidth}{\centering#3}%
  \parbox[#1]{.4\textwidth}{\raggedleft\texttt{#4}}\par\smallskip%
}

\makeatletter

\def\ps@IEEEtitlepagestyle{%
  \def\@oddfoot{\mycopyrightnotice}%
  \def\@evenfoot{}%
}
\def\mycopyrightnotice{%
 \textline[t]{\small \copyright 2024 IEEE.  Personal use of this material is permitted. Permission from IEEE must be obtained for all other uses, in any current or future media, including reprinting/republishing this material for advertising or promotional purposes, creating new collective works, for resale or redistribution to servers or lists, or reuse of any copyrighted component of this work in other works.}{\thepage}{}
  \gdef\mycopyrightnotice{}
}

\maketitle

\begin{abstract}
In a circular economy, products in their end-of-life phase should be either remanufactured or recycled. Both of these processes are crucial for sustainability and environmental conservation. However, manufacturers frequently do not support these processes enough in terms of not sharing relevant data about the products nor their (re-)manufacturing processes. This paper proposes to accompany each product with a digital twin technology, specifically the Product Digital Twin (PDT), which can carry information for facilitating and optimizing production and remanufacturing processes. This paper introduces a knowledge representation called Bi-Flow Product-Process-Resource Asset Network (Bi-PAN). Bi-PAN extends a well-proven Product-Process-Resource Asset Network (PAN) paradigm by integrating both assembly and disassembly workflows into a single information model. Such networks enable capturing relevant relationships across products, production resources, manufacturing processes, and specific production operations that have to be done in the manufacturing phase of a product. The proposed approach is demonstrated in a use-case of disassembling electric vehicle (EV) batteries. By utilizing PDTs with Bi-PAN knowledge models, challenges associated with disassembling of EV batteries can be solved flexibly and efficiently for various battery types, enhancing the sustainability of the EV battery life-cycle management.
\end{abstract}


\balance

\section{Introduction}

A circular economy is an economic system aiming at retaining resources in usage for as long as possible, realizing a closed-loop approach of reducing, reusing, remanufacturing, and recycling materials and products. It contrasts with the traditional linear economy, where resources are extracted, transformed into products, utilized, and thrown away as waste~\cite{cireconomy2017}. Recycling electric vehicle (EV) batteries is crucial for the circular economy because it can help reducing waste and conserve resources. EV batteries contain valuable materials such as lithium, nickel, cobalt or others depending on the specific battery chemistry, which can be recovered and reused. By recycling these materials, the need for mining new resources is reduced, which can have positive environmental and social impacts~\cite{Harper2019}. Conventional methods for EV battery recycling are based on crushing or shredding the entire battery or battery modules, separating the chemical elements, and reusing them for new batteries. On the contrary, another approach is based on reusing battery modules in a satisfactory state-of-health for stationary applications in households, industrial factories, energy sources, or energy storages~\cite{EVRecyclation}. This is the case that motivates the presented research. To be able to get battery modules in a good state, support for dismantling the entire EV battery is needed. Moreover, recycling or reusing EV batteries can also provide a solution for managing their end-of-life. As the count of EVs driving on the road is continually increasing, the number of batteries reaching their end-of-life will also increase. If these batteries are not properly managed, they can pose environmental and safety risks.

Production processes can be efficiently supported by a technology of digital twins, which pose digital replicas for physical systems~\cite{NOVAK2020, JAVAID2023}. In the domain of EV batteries, integration with digital twins is already a researched matter. Articles \cite{NASERI2023} and \cite{Panwar2021} provide extensive information on digital twins for EV batteries in several possible applications and usages as well as highlighting the benefits of using digital twins for remanufacturing and recycling EV batteries. This paper focuses on the application of digital twins in the end-of-life phase of EV batteries utilizing the so-called ``Product Digital Twin'' (PDT). PDTs improve sustainability and efficiency in product life-cycle~\cite{Pronost2024} management by offering manufacturers and remanufacturers a comprehensive tool for planning and optimizing the end-of-life processes of these products.

For representing PDTs, this paper proposes a concept of a Bi-Flow Product-Process-Resource Asset Network (Bi-PAN), building on the Product-Process-Resource Asset Network (PAN)~\cite{Biffl2021CBI}. PAN is a model that integrates product description, production processes, and resource information into a cohesive network/graph, modeling relevant relationships among these entities to support efficient planning, coordination, and optimization throughout production~\cite{Winkler2021ETFA}. PAN is designed in compliance with the industrial standard VDI/VDE~3682, which is widely accepted in the industry, especially in the automotive sector. Bi-PAN extends PAN by supporting both assembly and disassembly workflows within a single model, preserving the representation of products, processes, and resources along with their interconnections, while also incorporating additional features to handle reverse process flows. Thus, Bi-PAN supports the end-of-life phase of a product seamlessly together with production data. By integrating disassembly capabilities into the PAN, it serves to efficiently facilitate the design of remanufacturing and recycling processes in addition to the assembly process. The Product Digital Twin using the Bi-PAN model offers a holistic approach to product life-cycle optimization, facilitating the transition towards more sustainable and circular manufacturing practices.

\section{State of the Art}

The proposed approach is built on the top of two main research areas including (i)~Digital Twin and Product Life-Cycle, and (ii)~Product-Process-Resource Asset Network. These domains are described in detail in the following subsections.

\subsection{Digital Twin and Product Life-Cycle}

Digital Twin technology has widespread use in various industries and applications \cite{DT}. It is a digital replica or virtual representation of a physical object, system, or process. A digital twin can dynamically mirror and capture the characteristics of its real-world counterpart in real-time. It can also provide ways to simulate, analyze, maintain, and optimize the physical entity's operation and behavior in a virtual environment \cite{asi2021}. Moreover, defining a digital twin unambiguously can be challenging. Reference \cite{Emmert2023} provides a collection of various characterizations of digital twins, offering insights into the diverse interpretations.

Due to the fact that digital twin enables the storage and management of information and data about the products, including their design, materials, components, manufacturing processes, and maintenance history, digital twin is capable of product life-cycle management throughout the entire lifespan of the products. The key feature of a digital twin for its use in product life-cycle management is its nature as a continuously evolving entity, capable of providing ongoing value throughout its lifespan. They can be stored indefinitely and potentially offer valuable insights for future analysis, even after the physical entity has been disposed of \cite{JONES2020}.

Digital twin offers diverse applications, leading to the development of various frameworks and methodologies. These frameworks encompass a range of technologies, reflecting the versatility of digital twin creation. In the context of product life-cycle management, the approach proposed in reference \cite{Plociennik2022} highlights the significance of an Asset Administration Shell (AAS) \cite{AASPart1, AASPart2} as an option that can provide sustainable solutions choice while also ensuring interoperability. AAS, being a framework, requires an actual technology for digital twin creation within its parameters. AutomationML~\cite{Arndt2015}, for instance, offers an XML-based language tailored for describing automation systems. Utilizing AutomationML within the AAS framework~\cite{AutomationML2021} facilitates the creation of digital twins while adhering to its principles.

\subsection{Product-Process-Resource Asset Network}

The Product-Process-Resource Asset Network (PAN) \cite{Biffl2021CBI} is a graph-based formalism for expressing relationships among products, production processes, and resources. The PAN modeling is suitable for capturing and sharing knowledge among engineers and respective engineering tools as well as for identification of changes during engineering processes of multi-disciplinary system engineering projects. In fact, the PAN approach revolutionizes engineering processes by virtualizing assets, operations, and resources and especially by mapping and connecting them together by explicitly specified links. The usage of PAN optimizes the efficiency of engineering processes, it can support sustainability and enhances operations by modeling manufacturing processes, tracking resource usage, and monitoring asset performance. Through data integration, visualization, and analysis, PAN enables informed decision-making \cite{Winkler2021ETFA}. The PAN concept builds upon the Product-Process-Resource (PPR) \cite{AHMAD201833} concept/categorization, which consolidates product, process, and resource information into Engineering Assets. However, in the domain of Cyber-Physical Production Systems (CPPSs), these dependencies are frequently left implicit. The PAN addresses this gap by providing a structured model for explicitly articulating and overseeing the dependencies among product, process, and resource assets. We have been investigating possible extensions of the PAN paradigm by justifying the focus on a product structure, which has been addressed in~\cite{Strakosova2024}.

\section{Product Digital Twin and Bi-Flow Product-Process-Resource Asset Network}
\label{secPDTBiPAN}

This paper addresses the concept of the digital twin for a product, hereinafter called the Product Digital Twin (PDT).
The PDT represents a digital twin specifically designed for life-cycle management of products.
It facilitates easy access to relevant product information, including data from manufacturers such as maintenance and service details, operational data, or mechanical composition, as well as data gathered throughout the product’s life-cycle. This paper highlights the significance of information relevant for the end-of-life phase of products, as it represents a crucial opportunity to achieve sustainability. Specifically, the scope is focused on recycling and remanufacturing of products - both in terms of repair and re-commissioning and in terms of transformation into a new product. By leveraging data from manufacturers and accumulated life-cycle data, PDT aims at strengthening the efficiency and sustainability of remanufacturing and recycling processes. 

The benefits of the usage of the PDT in the end-of-life phase include:
\begin{enumerate}
    \item Designing remanufacturing/recycling process:
    PDT enables the design, planning, and simulation of the remanufacturing/recycling process before the physical product arrives. This pre-planning can significantly improve the efficiency of the remanufacturing/recycling process.
    \item Individualized approach:
    Each product having its own PDT allows for an individualized approach during the remanufacturing/recycling process. This is crucial because products may undergo changes or experience defects during their life-cycle. The PDT ensures that each product is handled appropriately based on its specific characteristic and condition.
    \item Real-time monitoring and adaptation:
    PDT provides online monitoring of the product's status during remanufacturing/recycling process. This capability is valuable for quick dealing with unexpected events or errors in the process.
    \item Data collection:
    The data collected during the remanufacturing and recycling process can be invaluable for manufacturers. They can analyze this data to gain insights into how products degrade or wear over time or to analyze problems during the remanufacturing/recycling process. This information can inform the design of the next generation of products, making them more sustainable. For remanufacturers/recyclers, this data could be used to make the remanufacturing/recycling process more efficient.
\end{enumerate}

While the advantages and applications of PDT are becoming increasingly clear, a significant challenge lies in creating a PDT that can effectively carry all the relevant information. When creating a PDT, it's crucial to clearly define the dependencies among individual components. This includes determining the sequence of component removal, understanding the method of removal (e.g., bolts, clips), and considering factors like accessibility and interconnectivity. These details are essential for automating the disassembly process. The PDT should be structured to represent the product's components and their relationships, rather than detailing specific disassembly procedures, as multiple disassembly strategies may exist. To address this need, the content of the Product Digital Twin (PDT) includes a generic disassembly plan. To visualize such a plan, a Bi-Flow Product-Process-Resource Asset Network (Bi-PAN) model is employed. This Bi-PAN model not only encompasses links between products and processes but also considers the resources required for each process. It facilitates the visualization of both assembly and disassembly processes, providing insights into the steps involved and the resources needed for efficient disassembly of the product.

\begin{figure}
    \centering
    \includegraphics[width=0.99\linewidth]{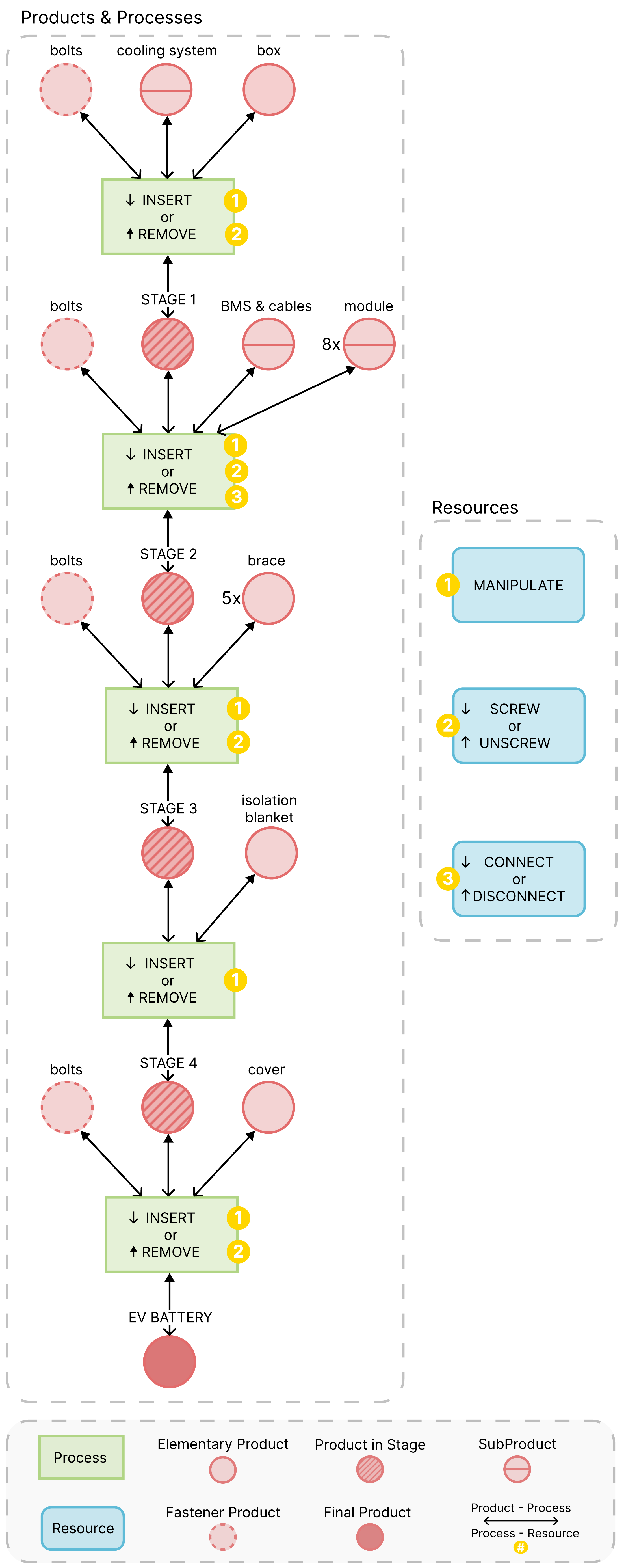}
    \caption{Bi-PAN for the simplified BMW i3 EV battery, representing product stages, production process operations, and resource skills. The legend explaining the nomenclature is included in the bottom part of the figure.}
    \label{figPAN}
\end{figure}

Just like the classic PAN, the Bi-PAN contains the same three basic elements (see Fig.~\ref{figPAN}): products as red circles, processes as green rectangles, and resources as blue rounded rectangles, all interconnected to represent their relationships. Black arrows show the connection between individual products and processes, forming an assembly or disassembly plan depending on the shape of the arrow. Open arrows represent assembly direction, while full arrows represent disassembly direction. Yellow markers link process steps with 
resources or resource skills, assigning specific assembly or disassembly procedures to individual processes. Depending on the direction of movement, directional arrows appropriately indicate the flow.
Within the Bi-PAN, products are categorized based on their assembly or disassembly role as follows:

\begin{enumerate}
    \item Elementary Product: A product that cannot be further disassembled.
    \item SubProduct: A product that is considered elementary in terms of decomposition within the extended PAN. It is composed of several parts and can be further disassembled.
    \item Fastener Product: This type of product serves as a fastener for other products.
    \item Final Product: The finished product, ready for use or sale.
    \item Product in Stage: An intermediate stage of the final product, representing a partially assembled or processed product within the production or disassembly process.
\end{enumerate}

The proposed approach is motivated by lessons-learned from the project dealing with automated disassembling of EV batteries. The EV battery is a simplified model of a BMW i3 battery~\cite{BMWi32018}, which is described in the next section.

\section{Product Digital Twin Based on the Bi-PAN for the Electric Vehicle Battery}

An EV battery serves as an exemplary case for demonstrating the PDT based on the Bi-PAN approach. Throughout the EV battery life-cycle, all three scenarios of the end-of-life phase are relevant: recycling to reintegrate scarce materials within the circular economy, repairing or replacing non-functional parts for re-use of EV batteries, and transforming EV batteries into modular units for secondary energy-storage applications. Remanufacturing or recycling of EV batteries also presents challenges, including the presence of hazardous substances and the risk of spontaneous combustion. It is therefore important to have relevant information and to select the appropriate remanufacturing/recycling route accurately and efficiently. This underscores the importance of PDT.

\begin{figure}
    \centering
    \includegraphics[width=0.75\linewidth]{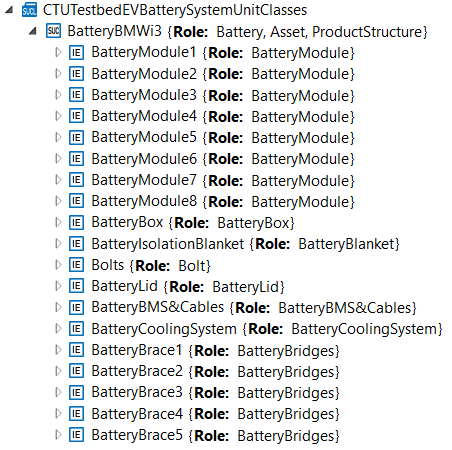}
    \caption{Example of a system unit class representing the BMW i3 battery in AutomationML as a screenshot from the AutomationML Editor.}
    \label{figAML}
\end{figure}

When it comes to selecting the appropriate recycling process for an EV battery, understanding the battery structure is paramount. This entails not just knowing the list of individual components but also comprehending the interdependencies between these components, including their composition. For the creation of the PDT, the utilization of the AutomationML tool proves invaluable. This tool enables not only the listing of individual components of the EV battery but also the depiction of their dependencies and detailed information. For instance, it can provide data about each component's position in the battery coordinate system, which proves beneficial when considering disassembly by a robot. An example of PDT and its structure in the AutomationML system unit class is depicted in Fig.~\ref{figAML}. The data obtained from AutomationML is important, but for the ability to plan assembly or disassembly operations, we lack the linking of the already described product structure with specific processes and resources. The Bi-PAN will serve us for this purpose.

As already introduced in Sec.~\ref{secPDTBiPAN}, Bi-PAN includes connections between products and processes, while also assigning the resources needed for each process. The Bi-PAN for the EV battery is depicted in the already mentioned Fig.~\ref{figPAN}. The following description explains the assembly process of an EV battery that starts from the top, and therefore the relevant direction for the black arrows is downwards. The assembly commences with three separate products: screws (Fastener Product), the cooling system (SubProduct), and the battery box (Elementary Product). These products are inserted into a single assembly through a process requiring resource skills for manipulation and screwing, as indicated by yellow markers. The output of this process is Stage 1 (i.e., Product in the Stage in the Bi-PAN notation) of the EV battery assembly. Stage 1 then proceeds to the next process along with bolts, BMS \& cables, and eight modules, where they are once again inserted into a single assembly. This process involves resource skills manipulation, screwing, and connecting cables. The result is Stage 2 of the EV battery assembly. Stage 2 and bolts, along with five braces, proceed to the next process where they are inserted into a single unit. This process requires resource skills for manipulation and screwing. The output of this process is Stage 3 of the EV battery assembly process. Stage 3, along with the isolation blanket, enters the subsequent process, requiring resource skills for manipulation only. This creates Stage 4 of the EV battery assembly. Finally, Stage 4, along with screws and a cover, enters the last process, requiring resource skills for manipulation and screwing. The output of this process is the Final Product - the assembled EV battery.

\begin{figure*}
    \centering
    \includegraphics[width=1\linewidth]{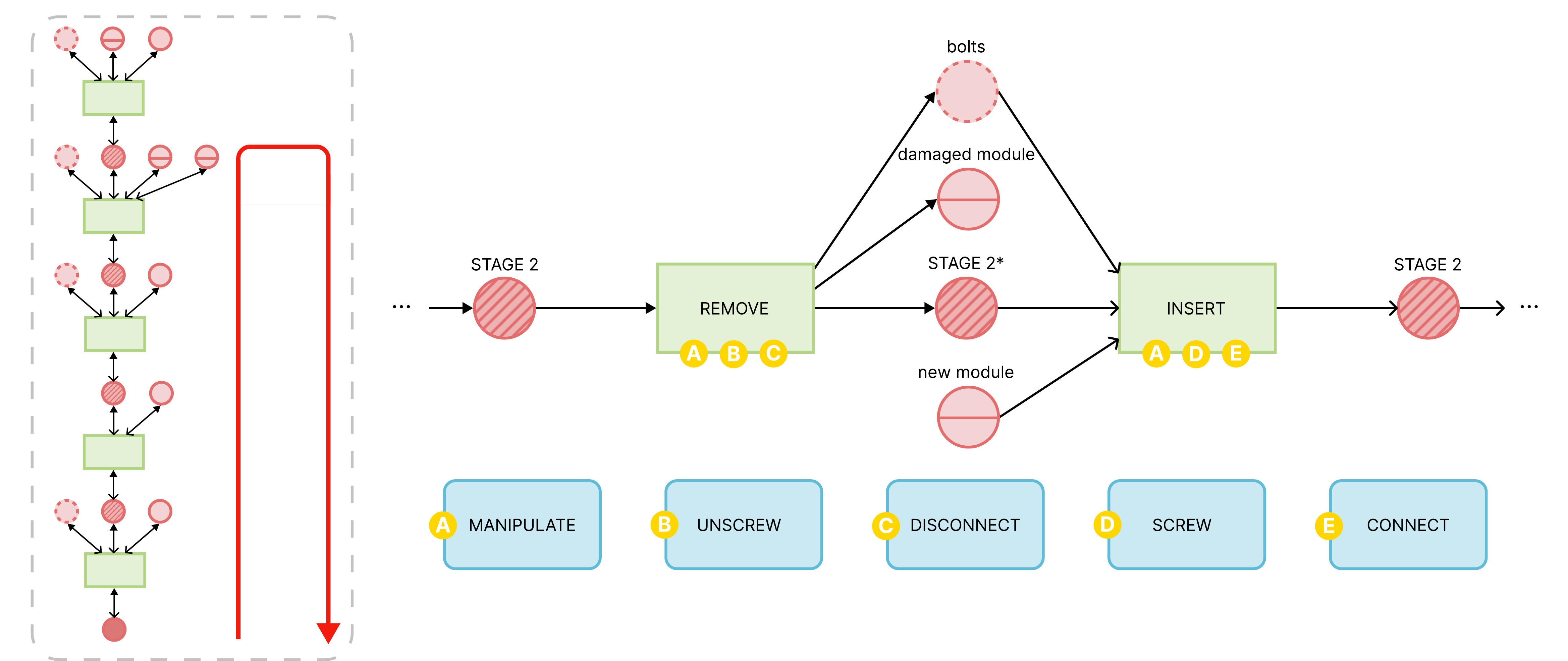}
    \caption{Changing a broken module use-case. In the generic Bi-PAN (depicted in Fig.~\ref{figPAN}), the module to be switched is found, the proper forward or backward operations are selected and the whole production recipe is extracted for this specific reparation case. Whereas the generic Bi-PAN has bidirectional arrows, this specific procedure extracted for the selected case has already single-directional arrows that can be already executed by the manufacturing system.}
    \label{figPAN2}
\end{figure*}

The aforementioned process can be reversed during remanufacturing or recycling of the product/battery. In such a case, individual product types are removed instead of being inserted, which from the resource skill viewpoint means that fastener products are unscrewed instead of screwed, cables/connectors are disconnected instead of connected, etc. This duality of Bi-PAN allows us to support the disassembly of the entire EV battery as well as some of its parts only.

To illustrate the application of the Bi-PAN in remanufacturing, we've selected the scenario of replacing a broken module with a new one. This specific process represents a sequence of operations that commences in a particular state of the EV battery and concludes at the desired state. The remanufacturing process depicted in Fig.~\ref{figPAN2} is derived from the generic Bi-PAN for the entire EV battery (i.e., from Bi-PAN in Fig.~\ref{figPAN}, which is repeated without labels on the left).
The flow of the whole remanufacturing process is depicted by a red arrow starting from the EV battery (Final Product) and continuing as a disassembly process to Stage 2, where the damaged module is removed and replaced by the new one, and it continues as an assembly process and ends as an EV battery (Final Product).

\section{Evaluation and Discussion}

During the research addressed in this paper, we have postulated the following research questions:

\begin{itemize}
    \item \emph{RQ-1: How can be the PAN paradigm extended to be applicable for remanufacturing and recycling processes?}
    \item \emph{RQ-2: What are the benefits of using the extended PAN approach as a basis for modeling DTs for EV batteries?} 
\end{itemize}

We have already used the (standard) PAN paradigm in the frame of other research projects and we found it to be efficient for capturing expert knowledge relevant for multi-disciplinary mechatronic production systems. However, the standard PAN does not provide support for remanufacturing or recycling processes. For this reason, we have extended the PAN paradigm in this paper and we have thus addressed the research question \emph{RQ1}. The proposed extension can express bi-directional manufacturing operations seamlessly in one graph, applicable (i) for manufacturing (i.e., forward direction through the Bi-PAN), and (ii) for remanufacturing/disassembling (i.e., backward direction through the Bi-PAN).

Addressing the research question~\emph{RQ-2}, the important benefit of the PAN paradigm (incl. the original one and the extended version proposed in this paper) is compliance with industrial standards such as ISA-95 or AutomationML. It can be used in conjunction with model-checking techniques to check the consistency of the overall PAN. The very same model in the Bi-PAN can be utilized for manufacturing and remanufacturing in intelligent factories. Added distinction between elementary/atomic products and (composite) products in PAN improves readability for humans but it is also useful for automated processing of PAN graphs.

\begin{figure}
    \centering
    \footnotesize
    \begin{tabular}{| c | c | c |}
        \hline
        & Conventional DT & Bi-PAN DT \\
        \hline
        \hline
        Product assembling & + + & + \\
        \hline
        Product disassembling & -- -- & + \\
        \hline
        Product remanufacturing & -- -- & + + \\
        \hline		
    \end{tabular}
    \caption{Evaluation of the digital twin design approaches based on Likert's scale: (i) the conventional DT design on the left and (ii) Bi-PAN-based DT design on the right.}
    \label{tabEval}
\end{figure}

Comparing and evaluating the proposed Bi-PAN approach for the digital twin design with the traditional DT design techniques is evaluated with Likert's scale in Fig.~\ref{tabEval}. The conventional design techniques are suitable for their original purpose only, it means for manufacturing/product assembling (see ``++'' in the upper left-hand side corner of Fig.~\ref{tabEval}). On the other hand, conventional design methods for DT do not provide enough support for product disassembling or remanufacturing, which are typically out-of-scope of these techniques (or that require a separate brand-new DT tailored for these purposes), see ``--'' in product disassembling and remanufacturing. The proposed Bi-PAN supports both assembling and disassembling of products (see ``+'' and ``++'' in the right-hand side column). A trade-off for supporting both is a slightly more complex model, requiring more modeling effort and time. This is the reason why the product assembling based on the Bi-PAN is evaluated with ``+'', compared to ``++'' in the case of the conventional DT design. The significant benefit of the Bi-PAN is the support for remanufacturing or reparation of products, as it is shown in the EV battery use-case in the previous section (``++''~on the Likert's scale in Fig.~\ref{tabEval}). With conventional DT design, support for disassembling is missing and the combination of product assembling and disassembling in the remanufacturing processes is not possible.

\section{Conclusion and Future Work}

Remanufacturing and recycling are key phases of the EV battery life-cycle in order to provide economic and ecologic sustainability for their usage. The approach proposed in this paper utilizes the product digital twin technology to persistently keep knowledge about the whole battery, and its modules, but also process steps to assemble, disassemble, or repair the battery. The proposed approach for the product digital twin design is based on the proposed extension of the product-process-resource asset network called Bi-PAN. 

The PDT based on Bi-PAN is used at the end-of-life phase of the battery. It enables querying of particular required processes including disassembling or repairing the EV battery which can be done either manually or on a flexible production system providing the required skills. 

Efficiency and sustainability pose key components for the future of remanufacturing and recycling processes and based on our experiments, the PDT concept based on Bi-PAN significantly contributes to these criteria.

In the \emph{future work}, we are going to focus on a better integration of the proposed PAN-based approach with the Asset Administration Shell for EV batteries.
We would also like to create a PAN model for a selected EV battery in full detail and discuss it with the car vendor of the EV, together with elaborating on the scalability and potential limitations of this approach on a real-life scale.

\section*{Acknowledgment}
This paper has been supported by the Grant Agency of the Czech Technical University in Prague, under grant No. SGS24/125/OHK2/3T/12 as well as by the project “Regeneration of used batteries from Electric Vehicles” (i.e., Slovak ITMS2014+ code 313012BUN5), the project is a part of the ``Important Project of Common European Interest'' (IPCEI) called the ``European Battery Innovation'' (i.e., code OPII-MH/DP/2021/9.5-34), which is a part of the ``Operational Program Integrated Infrastructure'' (EZOP No. 71235). The work was co-funded by the European Union (EU) under the project Robotics and advanced industrial production -- ROBOPROX (reg. no. CZ.02.01.01/00/22\_008/0004590).

\bibliographystyle{IEEEtran}
\bibliography{references}

\begin{thebibliography}{10}
\providecommand{\url}[1]{#1}
\csname url@samestyle\endcsname
\providecommand{\newblock}{\relax}
\providecommand{\bibinfo}[2]{#2}
\providecommand{\BIBentrySTDinterwordspacing}{\spaceskip=0pt\relax}
\providecommand{\BIBentryALTinterwordstretchfactor}{4}
\providecommand{\BIBentryALTinterwordspacing}{\spaceskip=\fontdimen2\font plus
\BIBentryALTinterwordstretchfactor\fontdimen3\font minus \fontdimen4\font\relax}
\providecommand{\BIBforeignlanguage}[2]{{%
\expandafter\ifx\csname l@#1\endcsname\relax
\typeout{** WARNING: IEEEtran.bst: No hyphenation pattern has been}%
\typeout{** loaded for the language `#1'. Using the pattern for}%
\typeout{** the default language instead.}%
\else
\language=\csname l@#1\endcsname
\fi
#2}}
\providecommand{\BIBdecl}{\relax}
\BIBdecl

\bibitem{cireconomy2017}
F.~Sariatli, ``Linear economy versus circular economy: A comparative and analyzer study for optimization of economy for sustainability,'' \emph{Visegrad Journal on Bioeconomy and Sustainable Development}, vol.~6, 2017.

\bibitem{Harper2019}
G.~Harper, R.~Sommerville, E.~Kendrick, L.~Driscoll, P.~Slater, R.~Stolkin, A.~Walton, P.~Christensen, O.~Heidrich, S.~Lambert, A.~Abbott, K.~Ryder, L.~Gaines, and P.~Anderson, ``Recycling lithium-ion batteries from electric vehicles,'' \emph{Nature}, vol. 575, pp. 75--86, 2019.

\bibitem{EVRecyclation}
A.~Beaudet, F.~Larouche, K.~Amouzegar, P.~Bouchard, and K.~Zaghib, ``Key challenges and opportunities for recycling electric vehicle battery materials,'' \emph{Sustainability}, vol.~12, no.~14, 2020.

\bibitem{NOVAK2020}
P.~Novák, J.~Vyskočil, and B.~Wally, ``The digital twin as a core component for industry 4.0 smart production planning,'' \emph{IFAC-PapersOnLine}, vol.~53, no.~2, pp. 10\,803--10\,809, 2020, 21st IFAC World Congress.

\bibitem{JAVAID2023}
M.~Javaid, A.~Haleem, and R.~Suman, ``Digital twin applications toward industry 4.0: A review,'' \emph{Cognitive Robotics}, vol.~3, pp. 71--92, 2023.

\bibitem{NASERI2023}
F.~Naseri, S.~Gil, C.~Barbu, E.~Cetkin, G.~Yarimca, A.~Jensen, P.~Larsen, and C.~Gomes, ``Digital twin of electric vehicle battery systems: Comprehensive review of the use cases, requirements, and platforms,'' \emph{Renewable and Sustainable Energy Reviews}, vol. 179, 2023.

\bibitem{Panwar2021}
N.~Panwar, S.~Singh, A.~Garg, A.~Gupta, and L.~Gao, ``Recent advancements in battery management system for li-ion batteries of electric vehicles: Future role of digital twin,'' \emph{Energy Technology}, vol.~9, no.~4, 2021.

\bibitem{Pronost2024}
G.~Pronost, F.~Mayer, M.~Camargo, and L.~Dupont, ``Digital twins along the product lifecycle: A systematic literature review of applications in manufacturing,'' \emph{Digital Twin}, vol.~3, p.~3, 2024.

\bibitem{Biffl2021CBI}
S.~Biffl, J.~Musil, A.~Musil, K.~Meixner, A.~Lüder, F.~Rinker, D.~Weyns, and D.~Winkler, ``An industry 4.0 asset-based coordination artifact for production systems engineering,'' in \emph{IEEE 23rd Conference on Business Informatics (CBI)}, vol.~01, 2021, pp. 92--101.

\bibitem{Winkler2021ETFA}
D.~Winkler, P.~Novák, K.~Meixner, J.~Vyskočil, F.~Rinker, and S.~Biffl, ``Product-process-resource asset networks as foundation for improving cpps engineering,'' in \emph{26th IEEE International Conference on Emerging Technologies and Factory Automation (ETFA)}, 2021.

\bibitem{DT}
F.~Tao, H.~Zhang, A.~Liu, and A.~Y.~C. Nee, ``Digital twin in industry: State-of-the-art,'' \emph{IEEE Transactions on Industrial Informatics}, vol.~15, no.~4, pp. 2405--2415, 2019.

\bibitem{asi2021}
M.~Singh, E.~Fuenmayor, E.~P. Hinchy, Y.~Qiao, N.~Murray, and D.~Devine, ``Digital twin: Origin to future,'' \emph{Applied System Innovation}, vol.~4, no.~2, 2021.

\bibitem{Emmert2023}
F.~Emmert-Streib, ``Defining a digital twin: A data science-based unification,'' \emph{Machine Learning and Knowledge Extraction}, vol.~5, pp. 1036--1054, 2023.

\bibitem{JONES2020}
D.~Jones, C.~Snider, A.~Nassehi, J.~Yon, and B.~Hicks, ``Characterising the digital twin: A systematic literature review,'' \emph{CIRP Journal of Manufacturing Science and Technology}, vol.~29, pp. 36--52, 2020.

\bibitem{Plociennik2022}
C.~Plociennik, M.~Pourjafarian, A.~Nazeri, W.~Windholz, S.~Knetsch, J.~Rickert, A.~Ciroth, A.~do~Carmo~{Precci Lopes}, T.~Hagedorn, M.~Vogelgesang, W.~Benner, A.~Gassmann, S.~Bergweiler, M.~Ruskowski, L.~Schebek, and A.~Weidenkaff, ``Towards a digital lifecycle passport for the circular economy,'' \emph{Procedia CIRP}, vol. 105, pp. 122--127, 2022, the 29th CIRP Conference on Life Cycle Engineering, April 4 – 6, 2022, Leuven, Belgium.

\bibitem{AASPart1}
``Details of the asset administration shell - part 1,'' Federal Ministry for Economic Affairs and Climate Action (BMWK), Plattform Industrie 4.0, Berlin, 2022, available online: \url{https://www.plattform-i40.de/IP/Redaktion/EN/Downloads/Publikation/Details_of_the_Asset_Administration_Shell_Part1_V3.html} [Accessed April 4, 2024].

\bibitem{AASPart2}
``Details of the asset administration shell - part 2,'' Federal Ministry for Economic Affairs and Climate Action (BMWK), Plattform Industrie 4.0, Berlin, 2021, available online: \url{https://www.plattform-i40.de/IP/Redaktion/EN/Downloads/Publikation/Details_of_the_Asset_Administration_Shell_Part2_V1.html} [Accessed April 4, 2024].

\bibitem{Arndt2015}
A.~Lüder and N.~Schmidt, \emph{AutomationML in a Nutshell}.\hskip 1em plus 0.5em minus 0.4em\relax AutomationML e.V. Office, 2015.

\bibitem{AutomationML2021}
``Interrelation of asset administration shell and automationml,'' 2021, available online: \url{https://www.automationml.org/wp-content/uploads/2021/07/PosPapier_Interrelation-of-AAS-and-AML.pdf} [Accessed April 24, 2024].

\bibitem{AHMAD201833}
M.~Ahmad, B.~R. Ferrer, B.~Ahmad, D.~Vera, J.~L. {Martinez Lastra}, and R.~Harrison, ``Knowledge-based {PPR} modelling for assembly automation,'' \emph{CIRP Journal of Manufacturing Science and Technology}, vol.~21, pp. 33--46, 2018.

\bibitem{Strakosova2024}
S.~Strakošová, P.~Novák, and P.~Kadera, ``Product-oriented product–process–resource asset network and its representation in {AutomationML} for asset administration shell,'' in \emph{29th IEEE International Conference on Emerging Technologies and Factory Automation (ETFA)}, to be published in 2024.

\bibitem{BMWi32018}
``Technical specifications. {BMW i3} (120 {Ah}),'' 2018, available online: \url{https://www.press.bmwgroup.com/global/article/detail/T0285608EN/technical-specifications-of-the-bmw-i3-120-ah-and-the-bmw-i3s-120-ah-valid-from-11/2018?language=en} [Accessed January 10, 2024].

\end{thebibliography}

\end{document}